\pdfoutput=1

\documentclass[11pt]{article}

\usepackage[]{acl}

\usepackage{times}
\usepackage{graphicx}
\usepackage{latexsym}

\usepackage[T1]{fontenc}

\usepackage[utf8]{inputenc}

\usepackage{microtype}

\newcommand{\name}{\texttt{DictaBERT}}

\title{\name: A State-of-the-Art BERT Suite for Modern Hebrew}

\author{Shaltiel Shmidman\textsuperscript{1,†}, Avi Shmidman\textsuperscript{1,2,‡}, Moshe Koppel\textsuperscript{1,2,†} \\
\textsuperscript{1}DICTA / Jerusalem, Israel \quad
\textsuperscript{2}Bar Ilan University / Ramat Gan, Israel \\ 
\texttt{\small \textsuperscript{†}\{shaltieltzion,moishk\}@gmail.com} \\
\texttt{\small \textsuperscript{‡}avi.shmidman@biu.ac.il}}

\begin{document}
\maketitle
\begin{abstract}
We present {\name}, a new state-of-the-art pre-trained BERT model for modern Hebrew, outperforming existing models on most benchmarks. Additionally, we release three fine-tuned versions of the model, designed to perform three specific foundational tasks in the analysis of Hebrew texts: prefix segmentation, morphological tagging and question answering. These fine-tuned models allow any developer to perform prefix segmentation, morphological tagging and question answering of a Hebrew input with a single call to a HuggingFace model, without the need to integrate any additional libraries or code. In this paper we describe the details of the training as well and the results on the different benchmarks. We release the models\footnote{Exact license can be found at \url{https://creativecommons.org/licenses/by-sa/4.0/}} to the community, along with sample code demonstrating their use. We release these models as part of our goal to help further research and development in Hebrew NLP. 
\end{abstract}

\section{Introduction}

In recent years, Hebrew NLP research has made major progress with the release of various pretrained language models with Hebrew support. The first was the multi-lingual transformer model mBERT, based on the BERT architecture \cite{devlin2019bert}, which was then followed by several other models of a similar architecture (HeBERT \cite{Chriqui_2022}, AlephBERT \cite{seker2021alephberta}, AlephBERTGimmel \cite{gueta2023large}, and HeRo \cite{shalumov2023hero}). Earlier this year it was shown that when finetuned, the much larger mT5 \cite{xue-etal-2021-mt5} models perform very well on the various Hebrew benchmarks \cite{eyal2022multilingual} resulting in SOTA scores for most of the experiments. 

We present our model {\name}  which is based on the BERT architecture with minor modifications to the training parameters, as well as improved training set and an improved procedure for preprocessing training samples. The model outperforms previous BERT models on most benchmarks, with the most noticeable improvement seen on the QA task. This gain is significant, as the QA task requires more complex syntactic understanding than other tasks. 
The performance gain on this task may indicate increased language comprehension in the model. Notably, the model's performance on the QA task is essentially equivalent\footnote{The QA task is scores with two measures: EM (exact match) and F1. {\name} achieves a slightly higher score for the EM, whereas mT5-XL achieves a slightly higher score for the F1.} to that of the mT5-XL model, a model with over 10 times as many parameters as {\name}. 

We also present \name\texttt{-morph}, \name\texttt{-seg} and \name\texttt{-heq}, 
 customized models fine-tuned for morphological tagging, prefix segmentation and question answering, respectively. The models are released with sample code for easy integration.\footnote{The sample code can be found in the appendix.} These fine-tuned models allow any developer to perform prefix segmentation and morphological tagging and question answering of a Hebrew input with a single call to a HuggingFace model, without the need to integrate any additional libraries or code. 
The details of the fine-tuning are detailed in section \ref{sec:finetuned-models}.

\section{Approach}

\subsection{Tokenizer}

We use the Word-Piece tokenization method proposed by \newcite{song2021fast} with the default normalizers and preprocessors suggested by HuggingFace, with the following modification:

We added a pre-tokenizer to handle usage of quotation marks and apostrophes in Modern Hebrew. We make sure to keep as single tokens words with quotation marks meant as abbreviations (e.g., \raisebox{-0.15\height}{\includegraphics[scale=0.35]{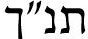}}) and words with an apostrophe indicating foreign sounds (e.g., \raisebox{-0.2\height}{\includegraphics[scale=0.35]{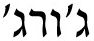}}). 

Following the work in AlephBertGimmel, the tokenizer was trained with a vocabulary size of 128,000 tokens. 

\subsection{Pre-training Data}

Our training dataset is a mixture of several sources, summing to a total of three billion words (3.8B tokens). We'd like to emphasize that none of the data is synthetic. The dataset is built up of several different components:

\textbf{C4 [70\%]}. We start with the HeDC4 corpus released by \cite{shalumov2023hero}, and continue further cleaning it. We removed approximately 15\% of the corpus using various techniques including histograms and gibberish detectors, as well as removing sentences that had a very high perplexity when run through an MLM scoring function. We limited our training corpus to contain mainly English and Hebrew; sentences that contained a disproportionate number of non-English/Hebrew tokens were removed. In addition, in order to retain only documents of substantial content, we filter out any document with less than 50 words. 

\textbf{Other sources [30\%]}. We collected data from various other sources including news sites, blogs, tv and movie subtitles, novels, and more. This data was also run through the cleaning process described above, although without the final filter by length. 

\subsection{Training Objectives}

In the original BERT article \cite{devlin2019bert}, the model was pretrained using two objectives: MLM (masked language modeling) and NSP (next sentence prediction). 
Following the work by \newcite{liu2019roberta}, we removed the NSP objective from the training, and only trained with the MLM objective. 

In addition, we made slight adjustments to the constructions of the training examples for the MLM objective, as follows: 

1. We don't mask tokens that are broken up into multiple word-pieces. In our research, we found that masking part of a word and training to predict it results in lower quality model. 

2. We never truncate part of a sentence. If we are creating a training example from a document that is longer than the maximum length, we remove entire sentences from the end of the example, such that we maximize the number of full sentences that can be fit within the training example, without any premature sentence truncation.

3. In order to help the model cope with data that isn't fully clean, and in order to expand the utility of the MLM function for downstream tasks, we insert randomly in 10\% of instances a [BLANK] token. This token helps the model learn that sometimes there can be a word in a sentence that doesn't actually belong there. An example of a downstream use case in which this would be useful is when utilizing the MLM feature to predict word alternates: the prediction of a [BLANK] indicates that the word can be optionally be left out altogether.

4. In addition, in order to keep the training data as clean as possible, we throw out any training instance with the [UNK] token, in order to ensure that we train on authentically Hebrew data. This resulted in throwing out approximately 3\% of the sentences in our already cleaned dataset. 

\subsection{Training Details and Hyperparameters}

We trained our model with the HuggingFace architecture wrapped with NVIDIA libraries\footnote{\url{https://github.com/NVIDIA/DeepLearningExamples/tree/master/PyTorch/LanguageModeling/BERT}} which are highly optimized for training compute-heavy machine learning models on NVIDIA hardware. We pre-trained the model on 4 A100 40GB GPUs for a total of 36,000 iterations, for a total of 35B tokens (1.75 epochs of each example 5 times). The training was done in 3 phases: 

(1) Sequences of up to 256 tokens with a learning rate of 6e-3 for 0.7 epochs. 

(2) Sequences of 256 tokens with a learning rate of 1e-4 for 0.3 epochs. 

(3) Sequences of 512 tokens with a learning rate of 5e-5 for 0.75 epochs. 

The total training time was 9.4 days. The training was done with a global batch size of 8,192, warmup proportion of 0.2843 for phase 1 \& 2, and 0.128 for phase 3. 

\section{Experiments and Results}

We compare the performance of {\name} with previous Hebrew models of similar paramater size, drawing results from their respective publications when they exist; otherwise we fill in the scores by finetuning the models ourselves. The models we compare to are: mBERT, AlephBert, AlephBertGimmel, HeRo, mT5-Base. For the QA task, we also compare our results to the much larger mT5-XL model (3.7B parameters).

Following previous publications, we tested our model on the following tasks:

\textbf{Morphology} We follow the multi-task configuration setup and evaluation procedure detailed in \newcite{seker2021alephberta} in order to fine-tune for segmentation, part of speech, and fine-grained morphological feature prediction, and in order to calculate the MultiSet (mset) scores for evaluation. We train the model with the same training \& test data used in previous works for this task. The results for this task are listed in Table \ref{tab:morph}. {\name} outperforms all previous models.

\textbf{Named Entity Recognition (NER)} We train on the NEMO dataset presented by \newcite{barekettsarfatinemo}, which contains 9 categories and 6,220 sentences (7,713 entities). We report the token F1 score for each model. The results for this task are listed in Table \ref{tab:sent-ner}. {\name} outperforms all previous models.

\textbf{Sentiment Analysis} We evaluate our model on the sentiment analysis dataset presented in \newcite{amram-etal-2018-representations}, based on 12K social media comments. The results for this task are also listed in Table \ref{tab:sent-ner}. {\name} outperforms all previous models.

\textbf{Question Answering} For this task we use the newly released HeQ dataset by \newcite{Cohen2023HeQ}, which contains 30K high quality samples from the GeekTime newsfeed, and from Wikipedia. The results for this task are listed in Table \ref{tab:qa}.  As discussed above, {\name} outperforms all previous models of similar parameter size, and performs essentially equivalent to the much larger mT5-XL model\footnote{We are also pleased to share the weights to our finetuned version of mT5-XL, after finetuning it on the HeQ corpus. The model is available for use on huggingface here: \url{https://huggingface.co/dicta-il/mt5-xl-heq}}.

\begin{table}[]
\centering
\begin{tabular}{|c|c|c|c|}
\hline
\textbf{}                & \textbf{Seg} & \multicolumn{1}{l|}{\textbf{POS}} & \textbf{Features} \\ \hline
\textbf{mBERT}           & 96.07             & 93.14                                  & 92.68                  \\ \hline
\textbf{AlephBert}       & 97.88             & 95.81                                  & 95.27                  \\ \hline
\textbf{ABG} & 98.09             & 96.22                                  & 95.76                  \\ \hline
\textbf{HeRo}            & 97.86             & 96.05                                  & 95.61                  \\ \hline
\textbf{mT5-Base}            & 96.34             & 95.9                                  & -                  \\ \hline
\textbf{\name}           & \textbf{98.16}             & \textbf{96.25}                                  & \textbf{96.1}          \\ \hline
\end{tabular}
\caption{Performance Comparison of Different Models on the Morphology task}
\label{tab:morph}
\end{table}

\begin{table}[]
\centering
\begin{tabular}{|c|c|c|}
\hline
\textbf{}                & \textbf{Sentiment} & \multicolumn{1}{l|}{\textbf{NER}} \\ \hline
\textbf{mBERT}           & 84.21              & 79.11                                    \\ \hline
\textbf{AlephBert}       & 89.02              & 83.62                                    \\ \hline
\textbf{AlephBertGimmel} & 89.51              & 85.26                                    \\ \hline
\textbf{HeRo}            & 89.56              & 86.79                                    \\ \hline
\textbf{mT5-Base}          & 87.25            & 74.7                        \\ \hline
\textbf{\name}           & \textbf{89.79}     & \textbf{87.01}                           \\ \hline
\end{tabular}
\caption{Performance Comparison of Different Models on the Sentiment and NER tasks}
\label{tab:sent-ner}
\end{table}

\begin{table}[]
\centering
\begin{tabular}{|c|c|c|}
\hline
\textbf{}                & \textbf{HeQ-EM} & \multicolumn{1}{l|}{\textbf{HeQ-F1}} \\ \hline
\textbf{mBERT}           & 62.7            & 71.42                                \\ \hline
\textbf{AlephBert}       & 57.91           & 67.66                                \\ \hline
\textbf{AlephBertGimmel} & 57.12           & 67.37                                \\ \hline
\textbf{HeRo}            & 61.89           & 71.3                                 \\ \hline
\textbf{mT5-Base}          & 53.65           & 65.03                        \\ \hline
\textbf{mT5-XL}          & 63.4            & \textbf{73.5}                        \\ \hline
\textbf{\name}           & \textbf{63.6}   & 72.9                                 \\ \hline
\textbf{ChatGPT}         & 9.38            & 32.19                                \\ \hline
\end{tabular}
\caption{Performance Comparison of Different Models on the QA task}
\label{tab:qa}
\end{table}

\section{Fine-tuned Models}
\label{sec:finetuned-models}

With this publication, we are also releasing three fine-tuned models, with sample code, in order to allow easy integration of DictaBERT's segmentation and morph analysis and question answering capabilities into any product or experiment. 

We wish to emphasize that for the morph analysis and segmentation models, although the evaluation above used the exact training and test data used in previous works in order to ensure fair comparison of DictaBERT's capabilities with previous BERT models, the fine-tuned models that we are currently releasing were fine-tuned on much larger corpora and with a simpler output format for easy integration, in order to provide maximum practical value to the community.

\subsection{\name\texttt{-seg}} 

This model was trained on a prefix segmentation task, wherein, given a sentence, we aim to identify the letters that function as proclitics at the beginnings of the words, segmenting them from the essential word unit. For example, the word \raisebox{-0.15\height}{\includegraphics[scale=0.32]{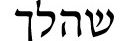}} ("that went") would be segmented into \raisebox{0.05\height}{\includegraphics[scale=0.32]{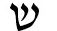}} ("that") and \raisebox{-0.15\height}{\includegraphics[scale=0.32]{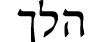}} ("went"). Note that this model does not segment suffixes, if any, at the ends of the words.

\textbf{Training Data} The training data for the fine-tuning of \name\texttt{-seg} is made up of various modern Hebrew texts including Wikipedia, blogs, and more. The prefix segmentation training set was initially automatically derived from Dicta's in-house diacritized corpus, and subsequently reviewed and corrected by an expert human annotator. The training data consists of 52K sentences with a total of 1.2M words. 

\textbf{Architecture} The architecture of the model is as follows: We run each word of the sentence through eight classifiers in order to predict the probability of each of eight different possible prefix functions a Hebrew proclitic may serve. We train the model to predict the relevant prefix classes for each word given its sentence context. During inference, we limit the predictions to valid sets of proclitic functions given the initial letters of the word. When words are comprised of multiple word pieces, we train on the first word piece alone, because the proclitics will virtually always be contained within the initial word piece (although we determine the valid sets of proclitic functions based on the whole word).

Sample code and output is displayed in Appendix \ref{sec:app-seg}. 

\subsection{\name\texttt{-morph}} 

This model was trained on a morphology task, where given a sentence, we aim to tag the fine-grained morphological features of each word. Specifically, \name\texttt{-morph} predicts part-of-speech, as well as gender, number, person, and tense, wherever relevant. Similarly to the previous model, this model also identifies proclitic functions. Additionally, this model also identifies whether there is a suffix appended to the word, and if so, which function it serves, and also the gender, number, and person of the suffix.

\textbf{Training Data} We used the UD Treebank \cite{sade-etal-2018-hebrew} which consists of 5K tagged sentences in the train split, as well an additional 35K sentences from the IAHLT\footnote{We would like to express our thanks to IAHLT for this tagged corpus. For more information regarding the resources curated and made available by IAHLT, see: \url{https://github.com/IAHLT/iahlt.github.io/blob/main/index.md}} UD corpus of tagged Hebrew sentences \cite{zeldes2022second}.

\textbf{Architecture} For every word in a sentence, we we train five classifiers to predict the morphological features of that word. For cases where a word is broken up into multiple word pieces, we perform the predictions on the first word piece. The 5 classifiers are as follows: 
\begin{itemize}
\item Prediction of the POS for the main word
\item Prediction of the proclitic functions (can be multiple or none - e.g., \raisebox{-0.15\height}{\includegraphics[scale=0.32]{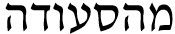}} ["from the meal"] has both an ADP and DET prefix)
\item Prediction of the fine-grained morphological features of the word (gender, number, person, and tense)
\item Prediction of whether there is a suffix and which function it serves
\item Predictions of fine-grained features of the suffix (gender, number, and person), if the previous classifier predicts that there is a suffix
\end{itemize}

Sample code and output is displayed in Appendix \ref{sec:app-morph}. 

\subsection{\name\texttt{-heq}} 

This model was trained on the question answering task, where given a context and a question, we aim to extract the answer to the question from the context. This model uses the standard HuggingFace architecture for Bert-Question-Answering. The model was trained on the HeQ train dataset as listed in the experiment section. 

Sample code and output is displayed in Appendix \ref{sec:app-heq}. 

\section{Conclusion}

We are happy to release these model to the public to help further research and development in Hebrew.\footnote{The base model {\name} is available at \url{https://huggingface.co/dicta-il/dictabert} \\ The fine-tuned segmentation model \name\texttt{-seg} is available at \url{https://huggingface.co/dicta-il/dictabert-seg} \\ The fine-tuned morphology model \name\texttt{-morph} is available at \url{https://huggingface.co/dicta-il/dictabert-morph} \\ The fine-tuned QA model \name\texttt{-heq} is available at \url{https://huggingface.co/dicta-il/dictabert-heq}.}

\bibliography{anthology}
\bibliographystyle{acl_natbib}

\clearpage
\onecolumn

\appendix

\section{Appendix: Example Code \& Usage of \name\texttt{-seg}}
\label{sec:app-seg}

\begin{figure}[hbt!] 
\includegraphics[page=1, scale=0.7]{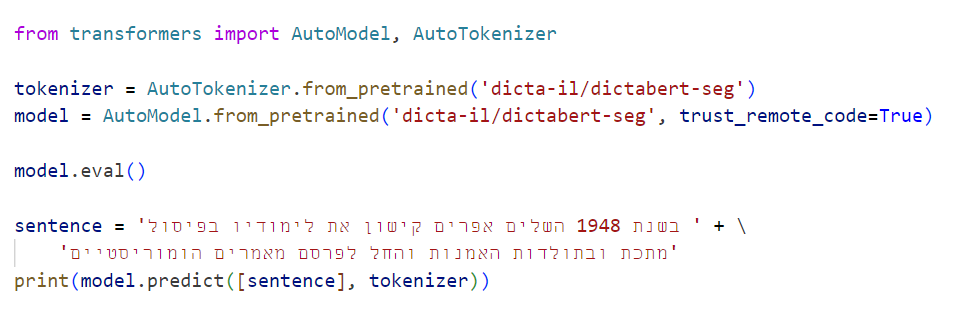}
\end{figure}

Output:

\begin{figure}[hbt!]
\includegraphics[page=1, scale=0.7]{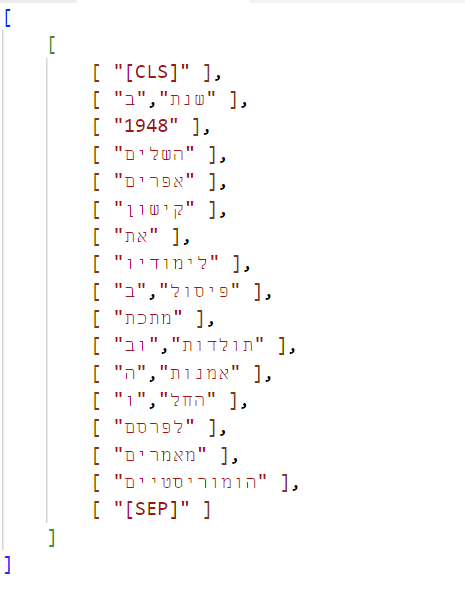}
\end{figure}

\clearpage 

\section{Appendix: Example Code \& Usage of \name\texttt{-morph}}
\label{sec:app-morph}

\begin{figure}[hbt!] 
\includegraphics[page=1, scale=0.7]{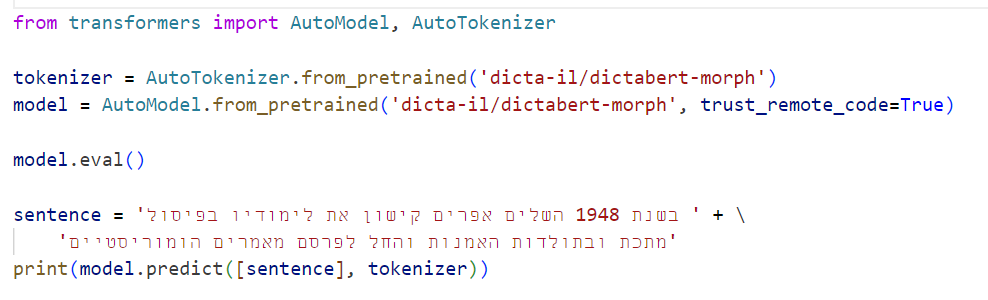}
\end{figure}

Output:

\begin{figure}[hbt!]
\includegraphics[page=1, scale=0.7]{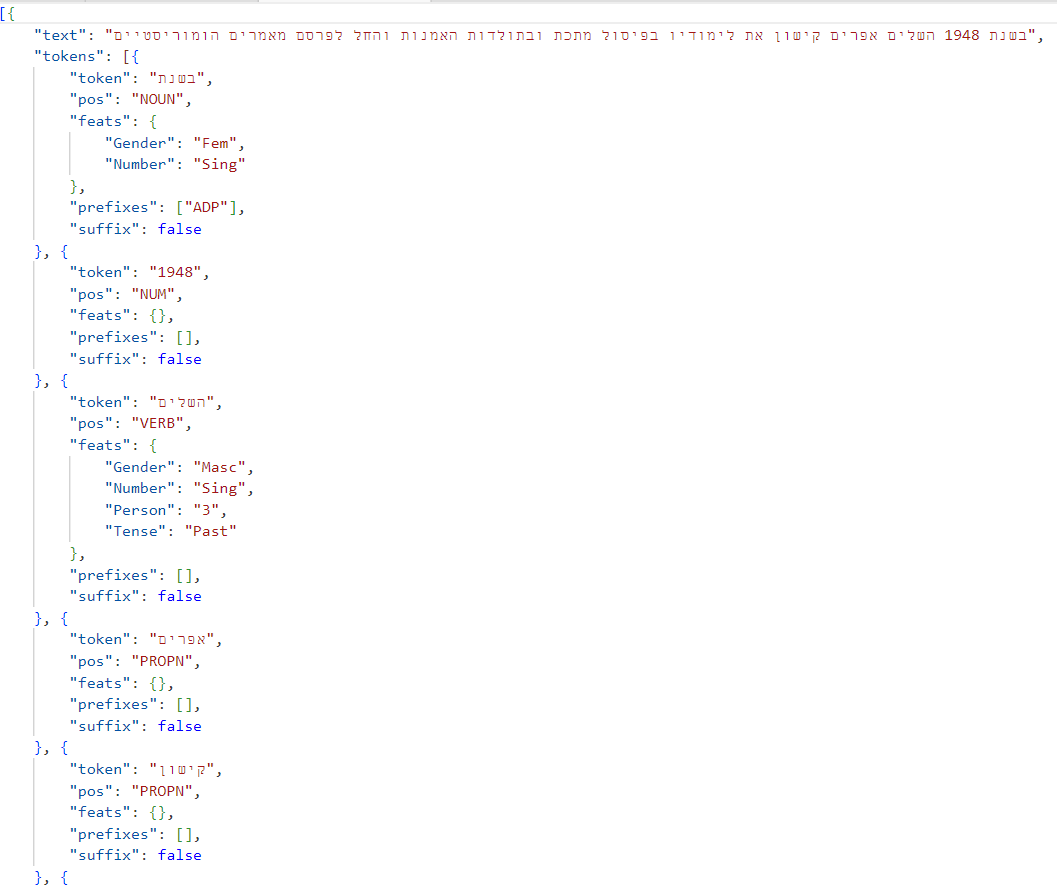}
\end{figure}

\begin{figure}[hbt!]
\includegraphics[page=1, scale=0.9]{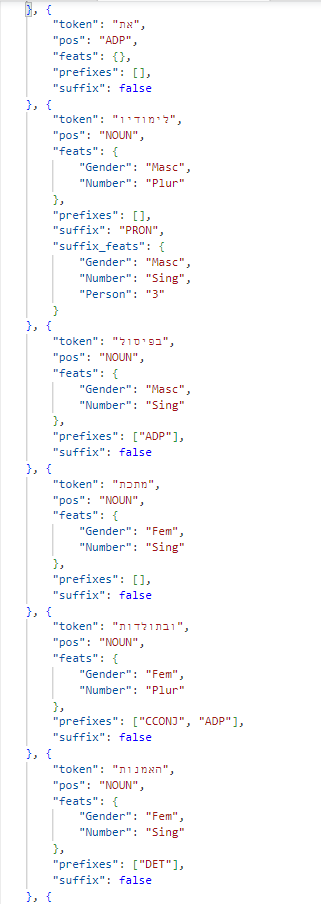}
\end{figure}

\begin{figure}[hbt!]
\includegraphics[page=1, scale=0.9]{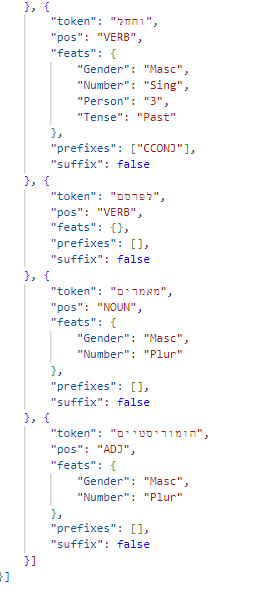}
\end{figure}

\clearpage

\section{Appendix: Example Code \& Usage of \name\texttt{-heq}}
\label{sec:app-heq}

\begin{figure}[hbt!] 
\includegraphics[page=1, scale=0.7]{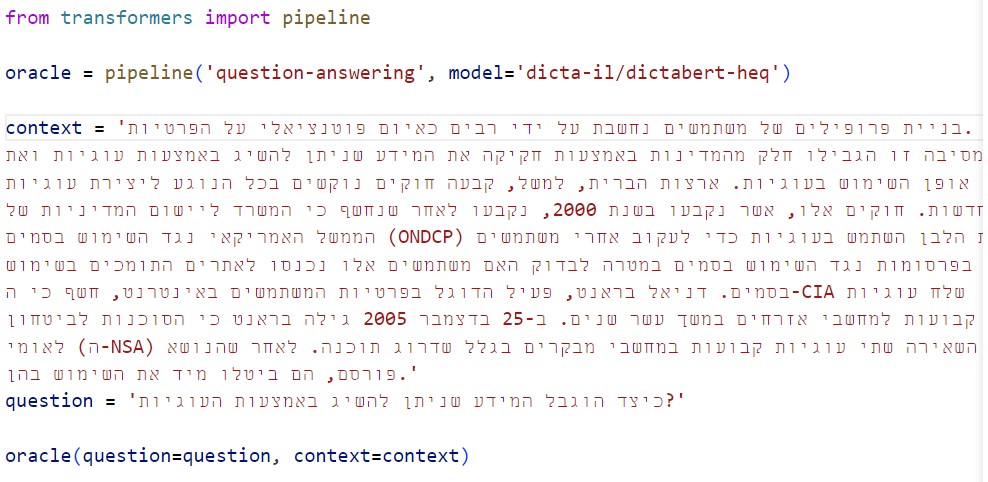}
\end{figure}

Output:

\begin{figure}[hbt!]
\includegraphics[page=1, scale=0.7]{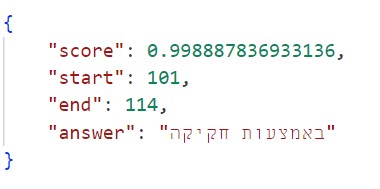}
\end{figure}

\end{document}